\documentclass[11pt,a4paper]{article}
\usepackage[hyperref]{acl2021}
\usepackage{times}
\usepackage{latexsym}

\usepackage{microtype}
\usepackage{enumitem}
\usepackage{times}
\usepackage{latexsym}
\usepackage{amsmath}
\usepackage{amsfonts}
\usepackage{url}
\usepackage{multirow}
\usepackage{comment}
\usepackage{graphicx}
\usepackage{booktabs}
\usepackage{lipsum}
\usepackage{booktabs}
\usepackage{siunitx}
\usepackage[T1]{fontenc}
\usepackage{multirow}
\usepackage{tabularx}
\usepackage{enumerate}% http://ctan.org/pkg/enumerate
\usepackage{colortbl}
\usepackage[utf8]{inputenc}
\definecolor{LightCyan}{rgb}{0.88,1,1}
\usepackage[english]{babel}
\definecolor{mycolor1}{rgb}{0.868, 0.188, 0.48}
\definecolor{bubblegum}{rgb}{0.99, 0.76, 0.8}
\definecolor{chocolate(traditional)}{rgb}{0.48, 0.25, 0.0}
\definecolor{bubbles}{rgb}{0.91, 1.0, 1.0}
\definecolor{darkblue}{rgb}{0.0, 0.0, 0.55}
\definecolor{orange}{rgb}{1,0.647,0}
\definecolor{darkpastelgreen}{rgb}{0.01, 0.75, 0.24}
\definecolor{buff}{rgb}{0.94, 0.86, 0.51}
\definecolor{blanchedalmond}{rgb}{1.0, 0.92, 0.8}
\definecolor{etonblue}{rgb}{0.59, 0.78, 0.64}
\definecolor{languidlavender}{rgb}{0.84, 0.79, 0.87}
\definecolor{junebud}{rgb}{0.74, 0.85, 0.34}
\definecolor{forestgreen(web)}{rgb}{0.13, 0.55, 0.13}
\definecolor{internationalorange}{rgb}{1.0, 0.31, 0.0}
\definecolor{mediumaquamarine}{rgb}{0.4, 0.8, 0.67}
\definecolor{lightmauve}{rgb}{0.86, 0.82, 1.0}
\definecolor{maize}{rgb}{0.98, 0.93, 0.37}
\definecolor{mygreen}{rgb}{0.13, 0.55, 0.13}
\newcommand\ourdataset{\textsc{DescGen}}

\aclfinalcopy 
\title{\ourdataset: A Distantly Supervised Dataset\\ for Generating Abstractive Entity Descriptions}

\author{Weijia Shi, Mandar Joshi, Luke Zettlemoyer\\
  Paul G. Allen School of Computer Science \& Engineering, \\ University of Washington, Seattle, WA \\
  \texttt{\{swj0419, mandar90, lsz\}@cs.washington.edu}
  }

\date{}
\begin{document}
\maketitle

\begin{abstract}

Short textual descriptions of entities provide summaries of their key attributes and have been shown to be useful sources of background knowledge for tasks such as entity linking and question answering.
However, generating entity descriptions, especially for new and long-tail entities, can be  challenging since relevant information is often scattered across multiple sources with varied content and style.
We introduce \ourdataset: given mentions spread over multiple documents, the goal is to generate an entity summary description. 
\ourdataset~consists of 37K entity descriptions from Wikipedia and Fandom, each paired with nine evidence documents on average. 
The documents were collected using a combination of entity linking and hyperlinks to the Wikipedia and Fandom entity pages, which together provide high quality distant supervision. The resulting summaries are more abstractive than those found in existing datasets, and provide a better proxy for the challenge of describing new and emerging entities. 
%Compared to other multi-document summarization tasks, our task is entity-centric, more abstractive, and covers a wide range of domains.
We also propose a two-stage extract-then-generate baseline and show that there exists a large gap (19.9\% in ROUGE-L) between state-of-the-art models and human performance, suggesting that the data will support significant future work.\footnote{Data and code available at \href{https://github.com/swj0419/DESCGEN}{github.com/swj0419/DESCGEN}
}
\end{abstract}
\section{Introduction}
% 1. entity knowledge is crucial for solving many tasks ....
% 2. recent work shows that it can be provided in textual form alone (entity knowledge comes from  textual description.)
% 3. curating knowledge is expensive..
% 4. propose the task
% 5. update new emerging entity

Entity knowledge has been shown to play an important role in various applications including language modeling~\cite{peters2019knowledge}, open-domain question answering~\cite{xu-etal-2016-question}, and dialogue generation~\cite{qin-etal-2019-entity}.
Recent studies suggest that such entity knowledge can be provided by simple textual descriptions~\cite{chen2019enteval}, which can be incorporated to improve downstream task performance~\cite{nie2018mention, logeswaran2019zero}.
However, manually curating entity descriptions is labor-intensive and it is challenging to keep pace with the ever growing emergence of new entities. In this paper, we present a new dataset \ourdataset~for automatically generating entity descriptions from relevant documents and mentions, which provides high quality supervision for a highly abstractive version of this task that targets early description of new entities as they emerge. For example, in Table~\ref{tbl:example}, machines are required to generate a description of \textit{Carl Menger}, given multiple documents mentioning him.  

\begin{table}[!tb]
\small
\begin{tabular}{|p{7cm}|}
\hline
\cellcolor[gray]{0.8} \textbf{Doc 1} \\ 
\hline 
...Are bitcoins, then, really worth anything? According to \textcolor{red}{\textbf{Carl Menger’s} subjective theory of value}, they are worth whatever individuals choose to believe they are worth. It is clear that many individuals value this new medium of exchange highly... \\ \hline
\cellcolor[gray]{0.8} \textbf{Doc 2}  \\ \hline
...The Austrian School of Economics has its roots outside of Austria — particularly in the French economists Jean Baptiste Say and Claude-Frederic Bastiat. \textcolor{forestgreen(web)}{The Austrian School proper began with \textbf{Carl Menger}}, who challenged the British labor theory of value. To learn more about Austrian Economics go to the website of The Ludwig von Mises Institute...\\ \hline
\cellcolor[gray]{0.8} \textbf{Doc 3}  \\ \hline
...Karl Menger was born on January 13, 1902, in Vienna. His father was \textcolor{blue}{the famous Austrian economist \textbf{Carl Menger} (1840–1921)} \textcolor{internationalorange}{who was one of the founders of marginal utility theory.}... \\ \hline
\hline
\cellcolor[gray]{0.8} \textbf{Entity Description}\\ \hline
\textcolor{blue}{\textbf{Carl Menger} (February 23, 1840 – February 26, 1921) was an Austrian economist} and \textcolor{forestgreen(web)}{the founder of the Austrian School of economics}. \textcolor{internationalorange}{He contributed to the development of the marginal utility theory} and to \textcolor{red}{the formulation of a subjective theory of value.} \\ \hline
\end{tabular}
\caption{An example from \ourdataset~exhibiting the diversity of source documents and the abstractive nature of the entity description summaries.}  \label{tbl:example}
\end{table}

%In support of the \ourdataset~task, w
\ourdataset~contains 37K entity descriptions extracted from Wikipedia and Fandom\footnote{Fandom is a set of encyclopedias centered around forms of entertainment such as movies, games etc.}. Fandom allows us to capture the key challenge of generating descriptions for emerging entities that are not in Wikipedia because they are less popular or have just been introduced to the public.
% We obtain source documents of entities by distant supervision method. 
To obtain source documents of the entities, we collect web documents and news articles where entity mentions are linked using web hyperlinks or an entity linker.
Our dataset is \emph{distantly supervised} in that these heuristically collected documents are not guaranteed to contain all the facts required to generate the description---as would be seen for natural text collections describing emerging entities. 
We also carefully annotate a subset of 1,000 examples to support more reliable evaluation (see Table~\ref{tbl:data_stats} for dataset statistics).
 
% Additionally,  
%\ourdataset~also poses a key challenge of identifying relavant information scattered in heterogeneous source documents and unifying it into a well-organized, typically highly abstractive description covering essential aspects of the entity. 
Unlike multi-document summarization that makes the assumption that a set of documents to be summarized are written on the same topic~\cite{zopf2016next}, \ourdataset~only assumes that source documents mention the entity. In contrast to an existing entity summarization benchmark~\cite[WikiSum]{liu2018generating}, \ourdataset~is more abstractive and better approximates challenges faced when describing new entities. Section~\ref{extractive} provides more details on these comparisons. Overall, our documents for generating a description can cover a much wider range of topics as well as text genres, including news, blog posts, and scientific articles.
For instance, the documents 1 and 2 mentioning \textit{Carl Menger} in Figure~\ref{tbl:example} discuss topics on bitcoins and the Austrian School of Economics. 
%The diversity in documents topics and text genres make more difficult to find entity-relevant information. 

%With a new task definition and its dataset, 
Finally, we also propose a two-stage method
that first extracts salient sentences relevant to the entity and then abstracts them into a description. We test a range of models to establish baseline results with both automatic and human evaluation. The best model based on BART~\cite{lewis-etal-2020-bart} achieves 28.2\% in the ROUGE-L F measure with a significant gap compared to the human performance 48.1\%, suggesting there was great room for future improvement. In summary, our contributions include:  
\begin{itemize}
    \item We propose a new dataset \ourdataset~that includes challenging, abstractive entity summaries. Our dataset contains over 37K pairs of entity descriptions and their associated documents, along with a human-annotated subset of 1,000 pairs.
    \item We conduct an extensive analysis of properties of the dataset and identify its challenges---extractive content selection from large amounts of text and abstractive generation from it, particularly for emerging entities.
    % including content selection and high abstractiveness. 
    \item We present a two-stage method and benchmark various models on our dataset, aiming to facilitate future work on this dataset.
    % demonstrating the difficulty of the task.
\end{itemize}

% a\begin{table}[hbt!]
% \centering
% \small
% {
% \vspace{-0.3cm}
% % \setlength{\tabcolsep}{0.3em}
%     \begin{tabular}{l|c|c|c|c|c}
%     \toprule[0.7pt]
%     \multirow{2}{*}{\textbf{Dataset}}   & \textbf{\# words}   & \textbf{\textbf{\# words}}  & \multirow{2}{*}{\textbf{\# examples}} \\
%     & \textbf{(docs)} & \textbf{(sum/desc)} \\  \hline
%     \ourdataset  & 17,623  & 49  & 122,872  \\    
%     Multi-News~\cite{fabbri2019multi} & 2,103 & 263 & 56,216  \\
%     DUC03+04~\cite{over2004introduction} & 4,636 & 109 & 320 \\
%     % CNN/DailyMail & 810 & 56 & 312,085  \\
%     WikiSum~\cite{liu2018generating} &  \\

%     \bottomrule[0.7pt]
%     \end{tabular} 
% }
% \caption{Comparison of our dataset with existing summarization datasets. \textbf{\# words (docs)} and \textbf{\# words (sum/desc)} are number of words in input documents and in the summary or description respectively.}
% \label{tbl:split}
% \end{table}

\begin{table}[!tb]
\centering
\small
{
    \begin{tabular}{l r r}
    \toprule[0.7pt]
     & Wikipedia & Fandom \\
     \hline
     {Entities} & 26,585 & 11,366\\
     {Documents} & 177,454 & 170,204\\
     {Input size} & 11,568 & 1,872\\
     {Output size} & 53 & 32\\
     {Human-authored descriptions} & 598 & 403 \\
    \bottomrule[0.7pt]
    \end{tabular} 
}
\caption{Basic statistics for \ourdataset. Input size and output size refer to the average number of words in the description and source documents respectively.}
\label{tbl:data_stats}
\end{table}

\section{Related work}
\label{sec:related-work}

\paragraph{Existing Entity Description Generation Task and Dataset} 
Previous works~\cite{novikova-etal-2017-e2e, cheng-etal-2020-ent, trisedya2020sentence} mainly take as input some structured data such as knowledge graphs to generate entity descriptions. However, knowledge graphs, often mined from text corpora, are overwhelmingly incomplete on real-world entities and may not be updated in real-time~\cite{dong2014knowledge}. Therefore, we focus on generating descriptions from natural language sources such as web texts and news because they are often primary sources for entities and have better coverage of entities across multiple domains.
\ourdataset~is most related to WikiSum, a recent dataset for generating Wikipedia summaries from textual sources~\cite{liu2018generating}. WikiSum source documents primarily come from high-quality articles cited in the Wikipedia pages which makes their data more extractive (Section~\ref{extractive}). 
In contrast, we collect our source documents heuristically using web texts and news, providing a better proxy for emerging entities where high-quality citation sources may not be available.
% it resembles use cases of generating descriptions for emerging entities when hand-picked citation sources are not available.
In addition, their evaluation is conducted only on distantly supervised test data. However, our experiments demonstrate 
%that the ranking list of models on distant supervised test set and 
that manually annotated data allows for much better evaluation of model performance (Table~\ref{result}).
%, highlighting the importance of carefully curated test data.

\paragraph{Multi-document summarization}  aims to condense a cluster of thematically-related documents into a short and informative summary. 
% There are two types of multi-document summarization tasks: generic summarization and query-focused summarization that summarizes objects specific to a query.
A wide range of multi-document summarization datasets have been built for the Document Understanding and Text Analysis Conferences~\cite{over2004introduction, owczarzak2011overview}, news~\cite{fabbri2019multi}, events~\cite{gholipour-ghalandari-etal-2020-large} and Wikipedia summaries~\cite{liu2018generating}. Recent work has studied both extractive~\cite{yasunaga-etal-2017-graph, nallapati2017summarunner,tohalino2018extractive} and abstractive summarization~\cite{banerjee2015multi, chali-etal-2017-towards, nayeem-etal-2018-abstractive}. However, existing datasets typically are not entity focused and assume the input documents are at least loosely centered around a coherent topic or event.

\paragraph{Wikipedia generation} Our work is also related to research on generating Wikipedia articles. For instance, \citet{sauper2009automatically} learn to build content templates using an integer linear program to generate full articles. Similarly, \citet{banerjee2016wikiwrite} generate Wikipedia pages by building a topic classifier to assign web retrieved contents into relevant sections.
We focus on a different task -- generating a short text description that can identify and best summarize an entity.

% \noindent \textbf{Pre-trained Language Generation Models.} 
\section{Dataset Collection}
\paragraph{Task definition}
Given a collection of documents $D = \{D_i | i = 1...n\}$ with mentions linked to the same entity $e$, 
% Given an entity $e$ and a collection of documents $D = \{D_i | i = 1...n\}$ that mention $e$,
the goal is to generate a description of $e$. For example, Table~\ref{tbl:example} shows a description of an entity (\emph{Carl Menger}) and three source documents with mentions.

\paragraph{Distant supervision}
We make use of existing knowledge bases, such as Wikipedia and Fandom, to collect entity descriptions.  
% We then pair these descriptions with source documents of entities automatically gathered from the Web to obtain our training data.
% To obtain source documents for entities, we leverage entity linker and 
To obtain source documents and mentions for each entity, we use a combination of hyperlinks to Wikipedia pages and an entity linker that links entity mentions in text.
Our dataset is \emph{distantly supervised} in that these heuristically collected documents are not guaranteed to contain all the facts required to generate the description. To analyze the quality of distant supervision, we collect a smaller \emph{verified} set of entity descriptions using human annotators.
In contrast with our work, WikiSum~\cite{liu2018generating} used documents cited in the Wikipedia pages or web pages returned by Google as source documents to generate Wikipedia lead sections. 
Because high-quality citation sources constitute a substantial part of overall documents (75\%), their dataset is less abstractive than \ourdataset~and unsuited for emerging entities where citations are not available. 
% Because our documents are heuristically collected, they are not guaranteed to provide enough signals to write a description. Therefore, we can view it as a form of distant supervision.
% based on the assumption that documents contain enough signals to generate a description. 
% For quality control, we filter out entites for which the unigram recall of the entity description against its concatenated source documents is lower than 0.6.
% Because the training data generated by distant supervision often contains varying amounts of noise, we apply an additional filtering process to remove low-quality examples: we compute the unigram recall of the entity description against its concatenated source documents, and filter out entities whose recall are lower than 0.6.

% concatenate source documents of an entity into a single mega-document and compute unigram recall between the entity description and the mega-document. We filter out entities who unigram recall was lower than 0.7. 

\paragraph{Sources}
% We constructed a new dataset to support \ourdataset~ described above. 
We paired entity descriptions with source documents from three sources: Wikilinks, RealNews, and Fandom using distant supervision. 
To capture the challenge of emerging entities, we retrieve source documents that are not in Wikipedia using Wikilinks and RealNews. We also include specialized entities in Fandom that do not have Wikipedia pages.
For quality control, we filter out entities for which the unigram recall of the entity description against its concatenated source documents is lower than 0.6.
% We gather the entity descriptions from Wikipedia and Fandom, and collect documents for entities from two sources: non-Wikipedia web pages and news articles. 

\subsection{Distantly supervised data collection}
\paragraph{Wikilinks} Wikilinks \cite{singh2012wikilinks} is a large dataset designed for cross-document coreference. It consists of non-Wikipedia web pages (discovered using the Google search index) containing entities that are hyperlinked to Wikipedia. For each entity, we retrieve a collection of web pages in Wikilink with the anchor text linked to it and use the lead section of target Wikipedia page as its description. We further parse the HTML texts of the web pages and extract contents as source documents. 

\paragraph{Real News} To expand the collection of source documents, we extract entity mentions in RealNews \cite{zellers2019defending}, a large corpus of news articles from Common Crawl. We first conduct a longest prefix match between the entity surface form and text tokens via trie, a prefix tree structure that supports efficient string searching. More specifically, we build a trie of entity names where each node is a word and its children indicate all possible continuations from the prefix. After retriving candidates for entity mentions, we use an off-the-shelf entity linking model \cite{GuptaSiRo17} to rank the candidates and add the corresponding news articles as source documents of the rank-1 candidate.

\paragraph{Fandom} Fandom\footnote{https://www.fandom.com/} is a collection of encyclopedias, centered around particular subjects and themes such as movies, TV shows, and games. It contains specialized entities that require domain experts with background knowledge to make edits. Entities and their source documents can be automatically extracted by internal links. We filter out entities and only keep those without Wikipedia pages, which can be viewed as new or emerging entities. The description of the entity is extracted from the lead section of its Fandom page. We collect data from the 32 largest Fandom Wikis.

% and availability.
% and categorize them into 5 domains: movie, game, TV series, fiction and cartoon. 

\subsection{Human-authored entity descriptions} \label{crowd}
Entity descriptions extracted from Wikipedia and Fandom have been authored and edited by multiple community contributors largely independently of our source documents. 
We collected additional entity descriptions via Upwork,\footnote{https://www.upwork.com/} a freelancing platform, to better analyze how descriptions sourced from documents in our dataset contrast with those from Wikipedia and Fandom. We provided the entity and its source documents to annotators on Upwork, and asked them to write the entity descriptions. The annotators are also asked to mark sentences they used to write the description. Each entity was assigned to 2 annotators. We collected 500 entity descriptions for dev examples and 500 descriptions for test examples. 

We control the quality of the crowdsourced descriptions by filtering annotators who produced low-quality descriptions. We ask every candidate to annotate the same 20 examples and use two criteria for narrowing down candidates:
(1) missing key information in descriptions 
(2) unjustified information in descriptions that cannot be inferred from source documents alone. 
Eventually, we filtered out 4 annotators and accepted 7 qualified annotators. The total annotation cost was around \$3500.

\subsection{Experimental setup}
All 37K entity description and document pairs in the dataset are randomly split into train, development and test sets. In addition to automatically collected descriptions from Wikipedia and Fandom, we use the human-authored descriptions (Section~\ref{crowd}) as verified subsets into dev and test splits. Table~\ref{tbl:split} shows basic statistics of the final dataset. 
We report model performance on automatically collected descriptions (distant) and human-authored descriptions (verified).  

\begin{table}[!tb]
\centering
\small
{
\setlength{\tabcolsep}{0.3em}
    \begin{tabular}{l|c|c|c|c}
    \toprule[0.7pt]
      \textbf{Entity source} & & \textbf{Train}   & \textbf{Dev}  & \textbf{Test}    \\  \hline
   \textbf{Wikipedia}  & Distant   & 21,267  & 2,659 & 2,659 \\
   \textbf{(Wikilinks + Real news)} & Verified & - & 299 & 299 \\
    \hline
   \multirow{2}{*}{\textbf{Fandom}}  & Distant  & 9,092  & 1,137 & 1,137 \\
   & Verified  & - & 202 & 201\\
    
    \bottomrule[0.7pt]
    \end{tabular} 
}
\caption{Number of entities for train, dev and test set.
% Basic statistics for our dataset. Distant and verified refer to automatically collected descriptions and crowdsourced descriptions respectively
}
\label{tbl:split}
\end{table}
The next section provides a detailed analysis of the data quality, including annotator agreement and other aggregate statistics.

\section{Dataset Analysis}
\label{sec:dataset-analysis}
An analysis of the data shows that \ourdataset~contains a high proportion of emerging entities from diverse domains, and is more extractive compared to other multi-document summarization datasets.

\begin{table}[!tb]
\centering
\small
{
% \vspace{-0.3cm}
% \setlength{\tabcolsep}{0.3em}
    \begin{tabular}{l|c|c|c|c}
    \toprule
    \textbf{Metrics}   & \texttt{R-1}  & \texttt{R-2}  & \texttt{R-L} & \texttt{METEOR}   \\ \hline
    IAA  & 45.8  & 36.1  & 47.7 & 23.3 \\    
    \bottomrule
    \end{tabular} 
}
\caption{Inter-annotator agreement (IAA) for human-authored descriptions.}
\label{tbl:agreement}
\end{table}

\subsection{Statistics} \label{stats}
% Table \ref{tbl:overview} shows statistics of our dataset compared with existing multi-document summarization datasets except for CNN/DailyMail. The total number of words of concatenated input documents in our dataset is much larger than in others. The large size of input text poses a significant challenge of extracting relevant information from very noisy text and makes training an end-to-end abstractive system infeasible. 
Table~\ref{tbl:data_stats} shows data statistics.
\ourdataset~contains about 37K entity descriptions from Wikipedia and Fandom. On average, each entity has nine source documents.  We can see that 36\% percent of entities come from Fandom, and therefore have never had a Wikipedia page written about them. 
%demonstrating that \ourdataset~ captures the challenge of emerging entities.

\begin{figure}
\centering
\includegraphics[width=0.8\linewidth]{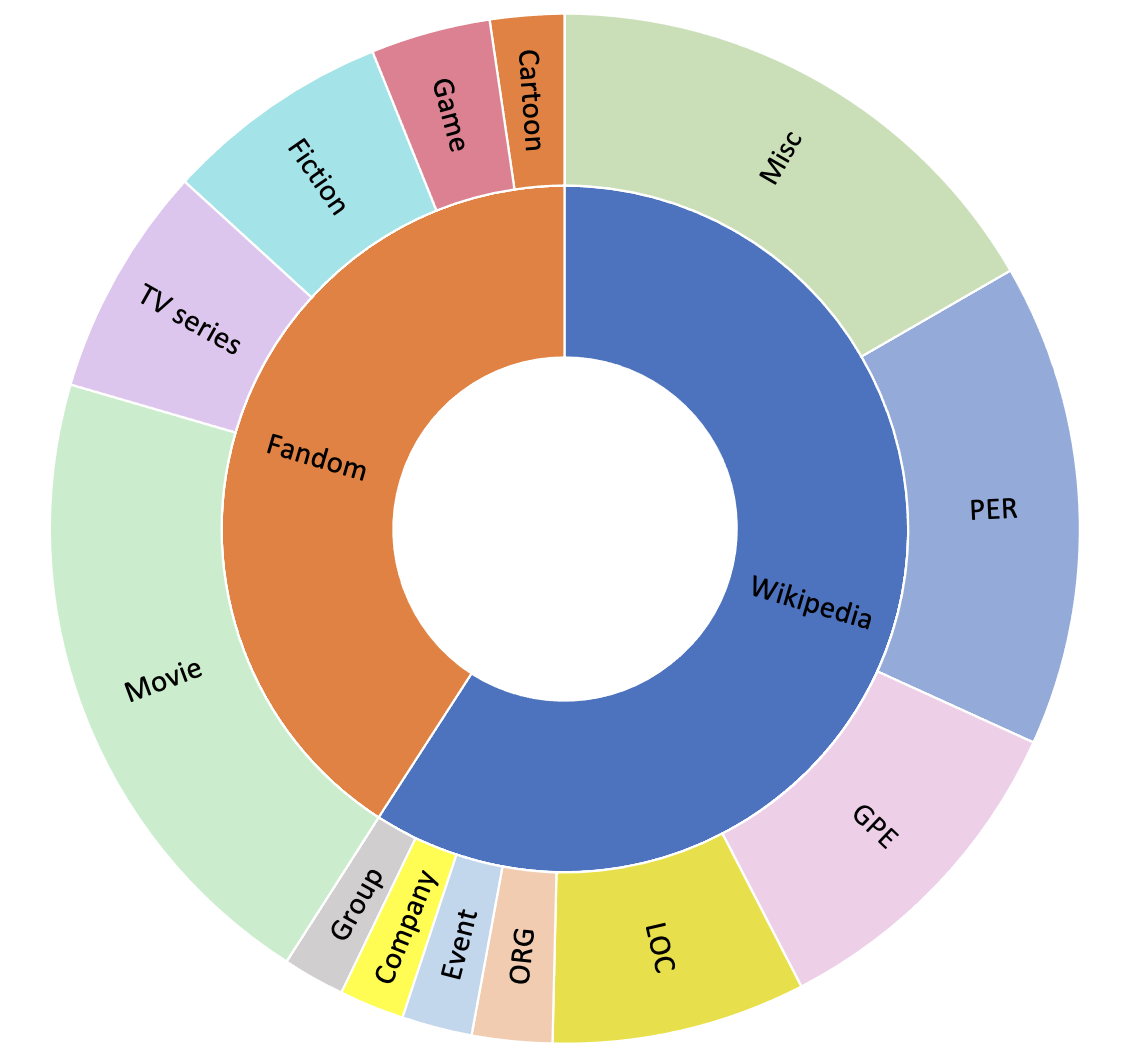}
\caption{
Distribution of entity domains (outer level) and knowledge sources (inner level).
}
\label{fig:diversity}
\end{figure}

\paragraph{Domain diversity} 
Figure \ref{fig:diversity} shows 
that \ourdataset~ covers a diverse set of entity domains.
For analysis, we associate entities in Wikipedia with domains (GPE, LOC, PER, ORG, EVENT, COMPANY, GROUP and MISC) by querying the DBPedia knowledge-base~\cite{lehmann2015dbpedia}. Each entity in Fandom is manually categorized into 5 domains: movie, game, fiction, TV series and cartoon based on its source Wiki. 
An analysis of baseline performance by entity type and domain (Section~\ref{sec:type_analysis}) reveals a notable drop for less popular domains such as Games and Fiction, highlighting generalization challenges.

% \noindent \textbf{Long-tail entities} 
% As shown in Figure \ref{fig:freq}, half of entities appear in fewer than 4 documents. These entities correspond to long-tail entities with relatively few mentions in text collections. 
% % entity popularity

% \begin{figure}[h]
% \centering
% \includegraphics[width=0.8\linewidth]{pic/freq.png}
% \caption{
% Histogram of the number of source documents per entity
% }
% \label{fig:freq}
% \end{figure}
% \vspace{-10mm}
\subsection{Inter-annotator agreement}
Each entity in the verified subset has two descriptions written by two annotators.
Following previous work~\cite{chen2015microsoft}, we quantify inter-annotator agreement on descriptions by treating one of the descriptions as the prediction and the other as the reference to compute ROUGE~\cite{lin2004rouge} and METEOR~\cite{denkowski2014meteor}. Table~\ref{tbl:agreement} shows high inter-annotator agreement of 47.7 in terms of ROUGE-L.

We additionally measure the agreement on content selection using sentences marked by annotators. In particular, agreement is achieved when both annotators selected the exact same sentences in all source documents for an entity. Cohen’s Kappa is 0.38, which indicates high agreement~\cite{brennan1981coefficient} considering the strict criterion of reaching agreement.

% We examine inter-annotator agreement on both content selection and descriptions. 
% The agreement between two annotators on sentences selected for writing descriptions are calculated in terms of Cohen’s K. 
% We examine inter-annotator agreement using the percentage of selected and non-selected sentenc
% 2. inter-annotator agreement
% - using the proportions of selected and non-selected units, Agreement is then calculated in terms of Cohen’s j (Radev et al. 2003) for two raters
% - rouge

% domain shift?

\subsection{Comparison between human-authored and Wikipedia/Fandom descriptions}
To understand how human-authored descriptions differ with Wikipedia and Fandom descriptions in terms of content and style, we compare them using automatic metrics (ROUGE) and manual evaluation.

\paragraph{ROUGE} 
Table~\ref{tbl:humanVSwiki} shows the averaged ROUGE scores of human-authored descriptions against Wikipedia and Fandom descriptions. 
Human-authored descriptions have higher word overlap with Wikipedia descriptions than with Fandom descriptions.  

\begin{table}[!tb]
\centering
\small
{
    \begin{tabular}{l|c|c|c}
    \toprule[0.7pt]
   & \texttt{R-1}  & \texttt{R-2}  & \texttt{R-L}   \\ \hline
    % All  & 34.1  & 18.6  & 33.8 \\   
    Wikipedia & 34.7 & 17.8 & 35.8 \\ 
    Fandom & 45.6 & 27.8 & 44.5 \\
    \bottomrule[0.7pt]
    \end{tabular} 
}
\caption{Rouge results on human reference against Wikipedia/Fandom descriptions.}
\label{tbl:humanVSwiki}
\end{table}

\paragraph{Pairwise comparison}
% We perform manual evaluation to compare if human-authored descriptions are qualitatively different from Wikipedia and Fandom descriptions.
Can humans distinguish between Wikipedia/Fandom and human-authored descriptions?
We have two human assessors evaluate 50 randomly sampled pairs of human-authored and Wikipedia/Fandom descriptions in a blind pairwise comparison, and ask them to classify descriptions into two categories: human-authored or Wikipedia/Fandom. The classification accuracy in Wikipedia and Fandom is 64.4\% and 61.1\% respectively and the inter-annotator agreement is 0.67 in Cohen’s Kappa. The relatively low classification accuracy suggests that there is no substantial quality and style difference in human-authored and Wikipedia/Fandom descriptions.
% no substantial quality difference

\paragraph{Quality analysis of distant supervision}
\begin{table}[!tb]
\centering
\small
{
\setlength{\tabcolsep}{0.4em}
    \begin{tabular}{l|c|c|c}
    \toprule[0.7pt]
    \textbf{Category}   & Paraphrasing &  Missing info. &  Extra details\\ \hline
    % All (count) & \\
    Wikipedia & 29 & 16 & 22 \\
    Fandom & 32 & 15 & 26 \\
    \bottomrule[0.7pt]
    \end{tabular} 
}
\caption{Number of times a human-authored description is classified into error categories with Wikipedia/Fandom descriptions as reference. The sample size is 40.}
\label{tbl:error}
\end{table}

% Having revealed that human reference descriptions are not qualitatively different from Wikipedia and Fandom descriptions, 
We are interested in understanding if automatically gathered documents can provide enough signals for writing the entity descriptions.
To study the quality of distant supervision, 
% we are interested in understanding why some human reference descriptions have low n-grams overlap (low ROUGE scores) with Wikipedia/Fandom descriptions. 
we manually analyze 40 human-authored descriptions that have low n-grams overlap with Wikipedia/Fandom descriptions, in terms of \textit{paraphrasing} (does the human-authored description express the same meaning but use different words?), \textit{missing information} (does the human-authored description miss any information in Wikipedia/Fandom description?) and \textit{extra details} (does the human-authored description contain extra details not included in the Wikpedia/Fandom description?). We use Wikipedia and Fandom descriptions as the ground truth and classify each human-authored description into one or more categories. The results are shown in Table~\ref{tbl:error}.
We find that the difference between the two sources of descriptions are mainly caused by paraphrasing and missing information. This suggests that even for entities that have very different human-authored and extracted descriptions, most of the information in the Wikipedia/Fandom descriptions is present in the documents.
% A relatively low proportion of errors are caused by missing information, suggesting that difference between human-authored descriptoins adn 
% paraphrasing and extra details, suggesting that missing information 
% This suggests that automatically gathered documents provide high-quality distant supervision for writing the entity descriptions.

% Wikipedia and Fandom descriptions are used as the ground truth.

% \textbf{Lingui}

% 1. Wiki summary and crowdsourced desc
% - manual classification: style similar
% - rouge
% \input{tables/humanVSwiki.tex}
% - manual evaluation: 
% missing details: 0 --> provide enough signal
% extra details
% paraphrasing

\subsection{Extraction vs abstraction}\label{extractive}
% \input{tables/gram}
% Our task can be approached by  is a process of extracting essential information from multiple documents about the entity and write a concise summary. 
% The abstractive and extractive method are two main approaches for summarization. 
% Entity description can vary in degree of 
Generating entity descriptions involves extracting essential information about the entity and condensing them into a short description.
To measure how much \ourdataset~requires paraphasing and compressing, we quantify the extractive nature of our dataset by the measuring extractive fragment coverage and density defined in
% We show in Table~\ref{tbl:recall} the n-gram recall between the summary and the input documents for multi-document summarization datasets. As the table shows, our dataset has low trigrams and 4-grams recall but high unigram recall, indicating our dataset is not likely to use novel words. 
% In addition, we quantify extractiveness by two measures defined by 
\citet{DBLP:conf/naacl/GruskyNA18}.
Extractive fragment coverage computes the percentage of words in summary that appear in source documents:
$$Coverage(A, S) = \frac{1}{|S|} \sum_{f \in F}{ |f| }$$
where $A$ is a concatenation of the source documents, $S$ is the description and $F$ is the set of shared token sequences in $A$ and $S$. Likewise, extractive fragment density is related to the average length of shared token sequences. 
For example, an entity description with high coverage and low density shares many individual words with source documents but almost no long phrases. 
$$Density(A, S) = \frac{1}{|S|} \sum_{f \in F}{ |f|^2 }$$

\begin{figure}[!tb]
\centering
\includegraphics[width=1\linewidth]{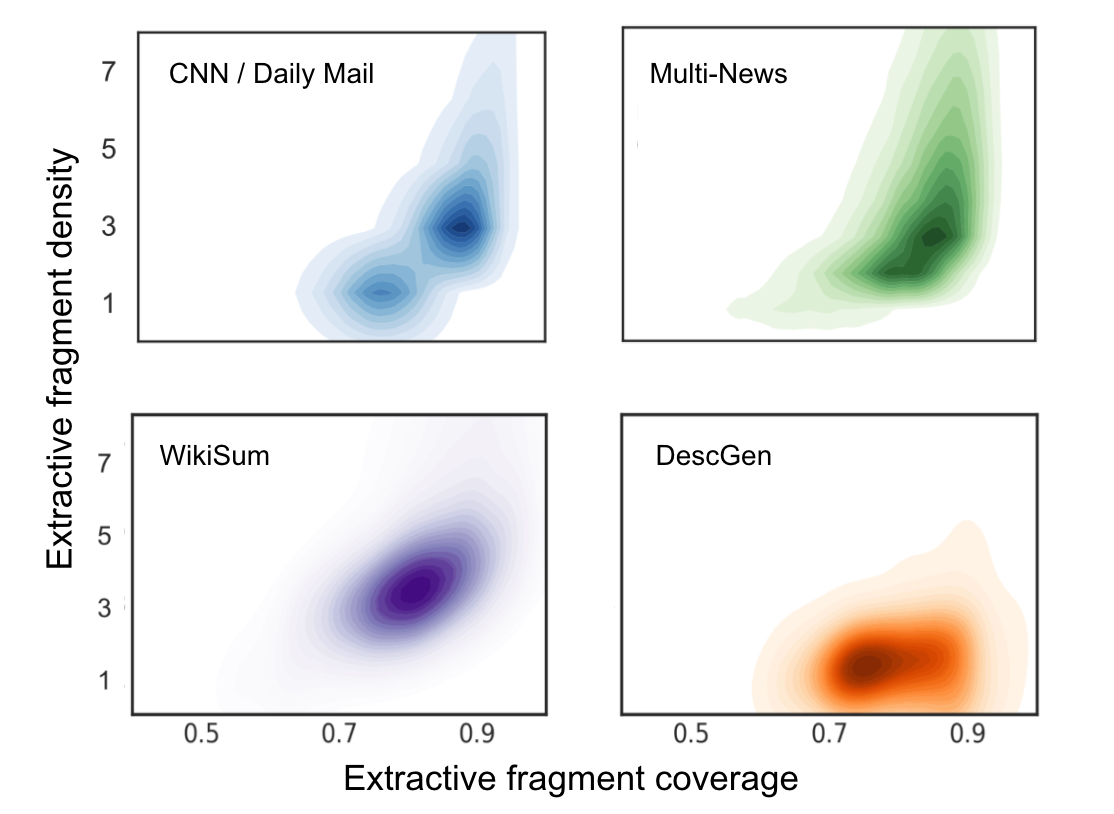}
\caption{
Density and coverage on different datasets. Large variability in y-axis reflects the variation in average length of shared token sequences. 
}
\label{fig:extrativeness}
\end{figure}

We compare our dataset with several multi-document summarization datasets, including CNN / Daily Mail, Multi-News~\cite{fabbri2019multi} and WikiSum~\cite{liu2018generating}. 
Figure \ref{fig:extrativeness} presents the density and coverage distribution. 
% Low density implies that copying of long sequences are not common. 
The density of Multi-News, CNN / Daily Mail and WikiSum are high, showing that there is much copying of long sequences with respect to source documents. \ourdataset~ shows high coverage but low density, suggesting it
is not common to copy long sequences and the data overall is much more abstractive.

%\ourdataset~ exhibits abstractive characteristics because it requires rearranging words to make meaningful sentences. 
% \ourdataset~ has lower density than Multi-News, CNN / Daily Mail and WikiSum. It suggests that

\section{Baselines}
In this section, we introduce several new baseline methods, building on state-of-the-art pre-trained models. The input documents can be long (Section \ref{tbl:recall}), making it computationally infeasible to train end-to-end models. 
We instead introduce a pipelined approach to generate an entity description in two stages. In the first extractive stage, a selector is used to identify representative sentences relevant to the entity from multiple source documents. In the second abstractive stage, a neural generation model is used to fuse the selected sentences to a description of the entity. We compare a number of different approaches for each stage, as summarized in the subsections below. 
% extractive only model

\subsection{Extractive stage}
 \begin{itemize}[label={},leftmargin=0pt]
\item \textbf{Trivial }concatenates all sentences that mention the entity, along with one sentence before and after each. The content is truncated to the first 1,000 tokens to fit the token limit of models in the abstractive stage.

\item \textbf{Cheating} ranks sentences according to their unigram recall against the description and selects the top 15 sentences. This heuristic demonstrates the effect of extraction on final performance.

\item \textbf{BERT}~\cite{devlin-etal-2019-bert} with a classifier uses a linear layer stacked on top of the BERT outputs and predict whether a sentence should be selected. The model is trained on our training dataset in which sentences are labeled by the cheating method.
% train by human marked data
\end{itemize}

\subsection{Abstractive stage}
We compare three pre-trained language generation models, including BART~\cite{lewis-etal-2020-bart}, T5~\cite{raffel2019exploring} and MARGE~\cite{lewis2020pre} to generate abstractive entity descriptions. We fine-tuned these models on our training dataset in a sequence-to-sequence fashion.  

\begin{itemize}[label={},leftmargin=0pt]
\item \textbf{T5}
is a text-to-text transformer pre-trained on a multi-task mixture of unsupervised and supervised tasks. 
% It shows a great advantage of scaling up model parameters.
We consider models of two sizes: base and large containing 220M and 770M parameters respectively. We use the Hugging Face version.\footnote{https://github.com/huggingface/transformers}

\item \textbf{BART} introduces a denoising autoencoder combining a bidirectional encoder and auto-regressive decoder. It is trained by reconstructing text corrupted with a noising function. We consider the base model with 139M parameters.
% base model with 6 layers and large model with 12 layers in encoder and decoder, released by Hugging Face. 

\item \textbf{MARGE} is a multi-lingual sequence-to-sequence model trained by reconstructing target documents retrieving paraphrased documents in other languages. It has around 960M parameters.
\end{itemize}

\section{Experiments}

\begin{table*}[htbp]
\small
  \begin{center}
  
    \begin{tabular}{c|c|ccc|ccc|ccc|ccc} % <-- Alignments: 1st column left, 2nd middle and 3rd right, with vertical lines in between
    \toprule[0.7pt]
      \multirow{3}{*}{\textbf{Extract.}} & \multirow{3}{*}{\textbf{Abstract.}} & \multicolumn{6}{c|}{Distant supervision} &  \multicolumn{6}{c}{Verified} \\

      & & \multicolumn{3}{c|}{Dev} &  \multicolumn{3}{c|}{Test} & \multicolumn{3}{c|}{Dev} &  \multicolumn{3}{c}{Test} \\
      & & \texttt{R-1} & \texttt{R-2} & \texttt{R-L} & \texttt{R-1} & \texttt{R-2} & \texttt{R-L} & \texttt{R-1} & \texttt{R-2} & \texttt{R-L} & \texttt{R-1} & \texttt{R-2} & \texttt{R-L} \\
      
      \hline
      \multirow{4}{*}{Trivial}  & BART & 24.5 & 11.3 & 22.1 & 23.4 & 10.7 & 23.8 & 27.2 & 14.2 & 26.4 & 27.1 & 15.9 & 26.6\\
      & T5-Base & 21.7 & 9.3 & 20.6 & 21.1 & 10.5 & 21.1 & 25.1 & 12.8 & 24.7 & 24.9 & 13.2 & 23.7 \\
      & T5-Large & 23.9 & 12.7 & 23.4 & 24.2 & 11.1 & 23.5 & 27.7 & 15.9 & 27.2 & 26.9 & 15.6 & 27.3 \\
      & MARGE & 23.2 & 10.6 & 21.8 & 23.0 & 10.2 & 22.1 & 26.4 & 13.9 & 25.8 & 26.2 & 14.0 & 25.8 \\

        \hline
      \multirow{4}{*}{BERT} & BART & \textbf{26.9} & 13.9 & \textbf{27.6} & \textbf{26.3} & 13.2 & \textbf{26.6} & \textbf{28.9} & \textbf{16.9} &  27.3 & 26.7 & \textbf{16.4} & \textbf{28.2} \\
    & T5-Base & 23.4 & 10.1 & 23.9 & 23.0 & 11.6 & 24.4 & 24.9 & 11.7 & 24.1 & 25.0 & 12.2 & 24.8  \\
    & T5-Large & 26.8 & \textbf{15.1} & 27.4 &   25.4 & \textbf{14.8} & 25.9 & 27.1 & 16.6 & \textbf{27.5} & \textbf{27.3} & 16.1 & 27.3 \\
      & MARGE & 25.1 & 13.8 & 26.2 & 24.9 & 11.9 & 25.0 & 26.7 & 15.7 & 25.9 & 26.3 & 14.8 & 25.8 \\
        \hline
    \multicolumn{2}{c|}{Human Performance}  & 40.7 & 21.9 & 39.9 &  39.1 & 21.8 & 39.3 & 45.2 & 36.7 & 48.7 & 45.3 & 35.4 & 48.1\\

         \bottomrule[0.7pt]

    \end{tabular}
      \caption{Experimental results of different baselines evaluated on distantly supervised and verified dev/test sets.}
    \label{result}
  \end{center}
\end{table*}

\subsection{Evaluation metrics}
% \paragraph{Automatic evaluation.} 
Following other summarization tasks, we evaluate the quality of generated descriptions by
ROUGE F1-score ~\cite{lin2004rouge}, which measures the overlap of unigram (R-1), bigram (R-2), and the longest matching sequence of words (R-L). 
In addition, we evaluate content selection by unigram and bigram recall to assess the importance of the extractive stage. Lastly, in addition to automatic evaluation, we also conduct human evaluation for non-redudancy, fluency, informativeness, and accuracy.

% to automatically evaluate the quality of generated descriptions. 

% To estimate human performance on the task, we treat one crowdsourced description as a system prediction and the other as a reference, and compute evaluation metrics.

% The results can also be used to quantify human agreement among annotators within each metric.

\subsection{Experimental results}
\label{sec:results}
\begin{table}[!tb]
\centering
\small
{
\setlength{\tabcolsep}{0.3em}
    \begin{tabular}{c|cc|cc|cc|cc} % <-- Alignments: 1st column left, 2nd middle and 3rd right, with vertical lines in between
    \toprule[0.7pt]
      \multirow{3}{*}{\textbf{Method}}  & \multicolumn{4}{c|}{Distant supervision} &  \multicolumn{4}{c}{Verified} \\

      &  \multicolumn{2}{c|}{Dev} &  \multicolumn{2}{c|}{Test} & \multicolumn{2}{c|}{Dev} &  \multicolumn{2}{c}{Test} \\
      & Uni. & Bi. & Uni. & Bi.  & Uni. & Bi.  & Uni. & Bi. \\
      
      \hline
      Trivial  & 60.5 & 23.8 & 59.9 & 23.4 & 78.8 & 50.4 & 76.9 & 43.2\\
    
        \hline
      BERT  & 65.1 & 26.1 & 66.9 & 27.7 & 80.4 & 50.6 & 77.5 & 43.8 \\
        \hline
      Cheating  & 72.4 & 31.7 & 72.3 & 31.4 & 81.6 & 51.9 & 79.2 & 44.6 \\
         \bottomrule[0.7pt]

    \end{tabular}
}
\caption{Unigram (Uni.) recall (\%) and bigram (Bi.) recall (\%) for extractive methods.}
\label{tbl:recall}
\end{table}
\begin{table}[!tb]
\centering
\small
{
    \begin{tabular}{l|ccc}
    \toprule[0.7pt]
    \textbf{Models}   &  BART & T5-Large & T5-base \\ \hline
    Non-redundancy & 3.8 & 3.5 & 3.6 \\
    Fluency & 4.6 & 4.7 & 4.6 \\
    Informativeness & 3.5 & 3.2 & 3.1\\
    Faithfulness & 2.7 & 2.5 & 2.6 \\
    \bottomrule[0.7pt]
    \end{tabular} 
}
\caption{Manual evaluation scores on a scale from 1 (very poor) to 5 (very high). All these models use BERT in the extractive stage.}
\label{tbl:manual_eval}
\end{table}

\paragraph{Automatic evaluation} In Table~\ref{tbl:recall}, we report the experimental results in the extractive stage. We observe that BERT consistently outperforms the unsupervised method Trivial, suggesting that training a model to predict sentence relevance can bring in immediate improvement in content selection. Meanwhile, the performance of BERT still lags behind the upper bound defined by Cheating by 1.7-7.3\% in unigram.
% , which brings interesting challenges for future work.

Table~\ref{result} presents ROUGE scores of various baselines in the abstractive stage. 
T5-large and BART show similar performance and outperform other models for both distant supervision and verified subsets, by a small margin. 
% Overall, BART with BERT as the extractive method achieves best performance 
Increasing model size from T5-base (220M) to T5-large (770M) parameters leads to a relatively large performance gain.  
The human baseline is superior to all the models and maintains a R-L score over 33 in distant supervision and 48 in the verified subset. The large gap between the human baseline and the best-performing model shows there is much room for future work.
 
\paragraph{Manual evaluation} We present two human assessors with source documents and descriptions generated from different abstractive models and asked them to rate descriptions in terms of \textit{non-redundancy} (does the description avoid repeating information?), \textit{fluency} (Is the description well-formed and gramatically correct?), \textit{informativeness} (does the description capture the salient information about the entity?) and \textit{faithfulness} (Is the description faithful to the source text?).  We compared BART, T5-Large, and T5-Base. For each model, we selected 100 descriptions and showed outputs of models to assessors side by side without revealing which model generates them. The score for each description was averaged between two assessors. As can be seen from Table~\ref{tbl:manual_eval}, BART shows strong performance on all dimensions, except for fluency. Overall, all three models can generate fluent descriptions (high \textit{fluency}) but struggle with producing accurate statements (low \textit{faithfulness}). In most cases of low faithfulness, we observe that the model directly copies words from the input that are not relevant to the entity as part of the description or synthesize information that are not directly inferable from the input.

% have poorly on ..., showing the pretrained language generation models tend to incorporate inaccurate details in generated descriptions. 
% the T-DMCA model does statistically significantly better on all dimensions, except on non-redundancy where tf-idf does about as well. Overall, we observed high fluency and coherence from our best abstractive model. Occasionally we observed some repetition of phrases which hurt the non-redundancy and structure, but it was much rarer compared with the other abstractive method, seq2seq.

\section{Analysis}
In this section, we perform qualitative and quantitative analysis of baseline results to better understand strengths and weaknesses of models, and hypothesize avenues for future work.

\subsection{Case study}
\begin{table}[]
\small
\begin{tabular}{|p{7cm}|}
\hline
\cellcolor[gray]{0.8} \textbf{Wikipedia description}\\ \hline
Carl Menger \textcolor{mygreen}{(February 23, 1840 – February 26, 1921)} was an Austrian economist and \textcolor{mygreen}{the founder of the Austrian School of economics}. He contributed to the development of the marginal utility theory and to \textcolor{mygreen}{the formulation of a subjective theory of value}. \\ \hline  
\cellcolor[gray]{0.8} \textbf{Human-authored description}\\ \hline
Carl Menger is an Austrian economist and one of founders of Marginal Utility Theory. \textcolor{mygreen}{He challenged the British labor theory of value and proposed subjective theory of value. He founded the Austrian School of Economics.} \\ \hline  \hline
\cellcolor[gray]{0.8} \textbf{BART} \\ 
\hline 
% \cellcolor{bubblegum} 
Carl Menger was an Austrian economist and one of the founders of marginal utility theory.
 \\ \hline
\cellcolor[gray]{0.8} \textbf{T5-Base}  \\ \hline
% \cellcolor{bubbles} 
Carl Menger was \textcolor{red}{born on January 13, 1902, in Vienna.} He was one of the founders of marginal utility theory. \\ \hline
\cellcolor[gray]{0.8} \textbf{T5-Large}  \\ \hline
% \cellcolor{blanchedalmond}
Carl Menger was an Austrian economist.   \\ \hline
\cellcolor[gray]{0.8} \textbf{MARGE}  \\ \hline
% \cellcolor{lightmauve} 
Carl Menger \textcolor{red}{(born January 13, 1902)} was an Austrian economist.  \\ \hline
\end{tabular}
\caption{Entity descriptions for \textit{Carl Menger} generated by different models. Red text indicates incorrect information in predictions while green text indicates information in the Wikipedia and human-authored descriptions that was not covered by any of the model predictions.}  \label{tbl:case}
\end{table}
A qualitative analysis of model predictions suggests that these models tend not to generate novel words in the description, and mostly copy words from the original text. The entity-centric nature of \ourdataset~makes extractive content selection difficult as evidenced by the gap between BERT extraction and the Cheating model (Section~\ref{sec:results}).
For example, Table \ref{tbl:case} shows the model-generated entity descriptions for \textit{Carl Menger} using source documents from Table~\ref{tbl:example}. BART, one of the best performing baselines, generates a description that has highest overlap with the Wikipedia description, but it still misses some important facts. T5-Base and MARGE confuse Carl Menger and his son, and incorrectly include information that does not describe the target entity. 
% Due to the abstractive nature of \ourdataset, future work can consider focusing on improving the level of abstraction in models.

\subsection{Entity knowledge in pre-trained models}
\begin{table}[!tb]
\centering
\small
{
    \begin{tabular}{l|cc|cc}
    \toprule[0.7pt]
    \multirow{2}{*}{\textbf{Models}}   &  \multicolumn{2}{c|}{Name-only} & \multicolumn{2}{c}{Regular}  \\ 
    & Fandom & Wiki. & Fandom & Wiki.\\ \hline
    BART & 12.7 & 16.6 & 27.5 & 28.4   \\
    T5-Base & 12.5 & 16.2 & 25.8 & 24.5 \\
    T5-Large & 11.7 & 16.8 & 26.1 & 27.6 \\
    \bottomrule[0.7pt]
    \end{tabular} 
}
\caption{Rouge-L scores for models evaluated on the verified test set. Name-only and regular refer to models using only the entity name as the input and models using source documents respectively. }
\label{tbl:entity_know}
\end{table}

BART, T5, and MARGE are language models pre-trained on text corpora including Wikipedia and Common Crawl. The parameters of the models appear to contain substantial linguistic and factual information~\cite{petroni-etal-2019-language, peters-etal-2018-dissecting}. 
% We are interested to know if the models can memorize entity descriptions in pretraining stage and if such additional entity knowledge improve model performance on generating entity descriptions.
% Therefore, we investigate the following questions: (a) How much entity knowledge do pre-trained language models contain? (b) Does additional entity knowledge improve model performance on generating entity descriptions?
In particular, we wonder if entity-related knowledge is captured in the pretraining stage and investigate the following questions: (a) Can the model memorize entity descriptions in pretraining stage?  (b) Does the memorized knowledge improve model performance on generating entity descriptions? 

To investigate the questions, we test the model's ability to write a description given only the entity name instead of source documents. 
% provide a prompt to the model and direct it to continue. 
We train the model on our training dataset to adapt to the style of Wikipedia in a similar way. The results are shown in Table~\ref{tbl:entity_know}.
Considering the name-only baselines, we can see that all of them perform worse on Fandom entities than Wikipedia entities. However, the regular baselines perform similarly on Fandom and Wikipedia. This result suggests that facts about entities learnt in pretraining stage have much less influence on  model performance when source documents are provided.
% entity-related knowledge is
% We can see that BART, T5-Base and T5-large can generate random information in BART but not , indicating that pre-trained language models can recall some facts  

% \subsection{Entity frequency}
% To analyze the impact of entity popularity in description generation, we evaluate performance a. 
% % As can be seen from Figure~\ref{asd},

% We analyze the performance of BART on the verified test set by entity frequency. As can be seen from Figure~\ref{asd}, the model performance is correlated with the entity frequency

% 1. entity frequency -- description length
%     - frequent ent: des long --> contain more details + content selection harder
%         - figure: x-freq, y-rouge score + extractive score + description length

% To check if entity knowledge , we i
% pre-trained models store more knowledge for popular entities 

\subsection{Entity type}
\label{sec:type_analysis}
\begin{table}[!tb]
\centering
\small
{
% \vspace{-0.3cm}
% \setlength{\tabcolsep}{0.3em}
    \begin{tabular}{lc|lc}
    \toprule[0.7pt]
    \textbf{Wikipedia}   &  ROUGE-L &  \textbf{Fandom} & ROUGE-L \\ \hline
    GPE & 28.6  & Movie & 28.1\\
    LOC & 28.5  & Game & 22.5\\
    PER & 23.7 & Fiction & 25.3\\
    ORG & 26.4 & Cartoons & 26.4\\
    Event & 25.6 & TV series & 27.6 \\
    Group & 20.2 &  \\
    Company & 21.4 & \\
    % Misc & \\
    \bottomrule[0.7pt]
    \end{tabular} 
}
\caption{ROUGE-L scores for BERT+BART evaluated on different entity domains in the verified test set.}
\label{tbl:type}
\end{table}

To understand how the performance of the models varies with different types of entities, we report the performance breakdown for different entity types in Table~\ref{tbl:type}. Among domains in Wikipedia, our model obtains low scores on group and company, suggesting that they are more challenging than other domains. In Fandom, entities from the game domain prove to be most difficult.

In summary, our analysis suggests there is room for improvement in extractive content selection and abstractive generation, particularly for new and emerging entities from less popular domains. 
% \subsection{Comparison with WikiSum}
% \input{tables/wikisum}
% We experiment with training models on WikiSum but testing it on \ourdataset. The results are presented in Table~\ref{tbl:wikisum}. For the model trained on WikiSum and tested on our Wikipedia test set, we observe a 3.1\% drop in performance compared with the model trained on \ourdataset. We attribute the low performance to domain differences between \ourdataset~and WikiSum.
% Considering the model trained on WikiSum and tested on our Fandom test set, we notice a larger drop in performance (4.7\%), showing the importance of including Fandom in the training set.
% % The significant gap between ... shows the importance of ....
\section{Conclusion}
In this work, we introduce \ourdataset, a new dataset for generating entity descriptions from mentions. \ourdataset ~contains 37K pairs of entity descriptions from Wikipedia and Fandom, and 481K automatically gathered source documents based on distant supervision. We also present a clean human-authored subset of 1,000 pairs for test. 
% We conduct extensive experiments to establish baseline results on our dataset. 
We show that, as compared to existing benchmarks, \ourdataset~requires more abstractive summaries, which we argue better approximate the challenge of describing emerging entities. We also show that the performance of state-of-art models is far from human levels, suggesting that our task remains a significant challenge with room for improvement.
Our study points to an interesting research direction on modeling entity knowledge from contexts. We hope it will facilitate future work on incorporating entity knowledge into downstream tasks and generating descriptions for emerging entities. 

% \section{Acknowledgement}

\section*{Acknowledgements}
This work was supported in part by the ARO (AROW911NF-16-1-0121) and the NSF (IIS1562364). The authors would like to thank Ari Holtzman, Bhargavi Paranjape, Elizabeth Clark, Terra Blevins and anonymous reviewers for helpful comments. 

\bibliographystyle{acl_natbib}
\bibliography{anthology,acl2021}

\clearpage
\appendix
\section{Appendix}
\subsection{Experimental Details}
All the abstractive models are initialized from the pretrained models. The BART, T5-base and T5-large are adopted by the huggingface framework~\cite{wolf2020transformers}. The MARGE model is adopted by the official authors~\cite{lewis2020pre}. We apply the Adam optimizer~\cite{kingma2014adam} with $\beta_1= 0.9, \beta_2=0.999, \epsilon = 1e-08$. The learning rate is selected from \{1e-3, 0.5e-3, 1e-4, 0.5e-4, 1e-5, 0.5e-5\}. The best learning rate for BART, T5-base, T5-large and MARGE is 1e-5, 1e-5, 0.5e-5,0.5e-4.
We use beam searching with beam-size 5 as decoding algorithm, which is selected from \{5, 10, 15, 20\}. We use the batch size of 5 for all models due to memory limit. 

\subsection{More examples}
See next page.
\begin{table*}[t]
\small
\begin{tabular}{|p{15.3cm}|}
\hline
\cellcolor[gray]{0.8} \textbf{Doc 1} \\ 
\hline 
...It sometimes gets confusing in the global village , where technology, finance, cross-cultural interactions, and expanding ethnic diasporas are tearing apart the relationship between borders and making multiple identities possible. Hence, Ang Lee is a Taiwanese artist who directs American films, but he is also an American film director of Chinese movies. As a member of the Sinosphere, enlarged by fifty million overseas Chinese, Ang is not only a creative individual who makes our world more interesting and prosperous. He also helps to bridge between nations and cultures and to produce a Sino-American synergy that is more conducive to peace than a contingency of Chinese and U.S. diplomats...  \\ \hline
\cellcolor[gray]{0.8} \textbf{Doc 2}  \\ \hline
...The Life of Pi. One of the most interesting film adaptations set for release in 2012 is Brokeback Mountain fame. Suraj Sharma, who has no previous acting experience, will play the central character, Piscine Patel. Based on the novel by Yann Martel, it is being brought to the big screen by Ang Lee... \\ \hline
\cellcolor[gray]{0.8} \textbf{Doc 3}  \\ \hline
Comic character Hulk is Dr. Bruce Banner, who becomes a green monster with powerful strength after an experiment went bad, or well, depending on who you ask. In 2003, director Ang Lee's film Hulk brought this character to the big screen, but was poorly received by Hulk's fans...\\
\hline
\cellcolor[gray]{0.8} \textbf{Wikipedia Description}\\ \hline
Ang Lee, (born October 23, 1954, P’ing-tung county, Taiwan), is an Taiwan-born film director \textcolor{mygreen}{who transitioned from directing Chinese films to major English-language productions.}\\ \hline
\cellcolor[gray]{0.8} \textbf{Human-authored Description}\\ \hline
Ang Lee is a Taiwanese director \textcolor{mygreen}{who directs American and Chinese films. He is a director of the Life of Pi and Hulk and regarded as Second New Wave of Taiwanese directors.}\\ \hline
\cellcolor[gray]{0.8} \textbf{BART}\\ \hline
Ang Lee is a Taiwanese film director and screenwriter.\\\hline
\cellcolor[gray]{0.8} \textbf{T5-base}\\ \hline
Ang Lee is a Taiwanese film director. \\\hline
\cellcolor[gray]{0.8} \textbf{T5-large}\\ \hline
Ang Lee is a Taiwanese film directors and screenwriter \\ \hline
\end{tabular}
\end{table*}
\\~
\hspace{2em}
\begin{table*}[t]
\small
\begin{tabular}{|p{15.3cm}|}
\hline
\cellcolor[gray]{0.8} \textbf{Doc 1} \\ 
\hline 
...In the summer of 1994, Arthur managed to get himself and his family (as well as Harry and Hermione) tickets for the 1994 Quidditch World Cup from Ludovic Bagman because Arthur had helped Otto Bagman, Ludo's brother, out of a minor scrape. Arthur was among the Weasleys who fetched Harry from the Dursley family via the Floo Network. While there, he expressed his fascination at various Muggle artefacts in the Dursley's house.The group sat in the Top Box, where they were confronted by the Malfoy family, who were there by a personal invitation from the Minister himself, though both Arthur and Lucius were able to restrain themselves out of respect for Cornelius Fudge...\\\hline
\cellcolor[gray]{0.8} \textbf{Doc 2}  \\ \hline
...Before working at the Ministry, he was a Beater for both the Wimbourne Wasps and the English National Quidditch team. He had a brother named Otto Bagman. He also tended to play dirty when gambling and betting as he tried to find loopholes or even pay in fake money/gold... \\ \hline
\cellcolor[gray]{0.8} \textbf{Doc 3}  \\ \hline
...A lawn mower is found in the Muggle Studies classroom at Hogwarts School of Witchcraft and Wizardry. Arthur once helped Ludovic Bagman's brother, Otto Bagman, by smoothing over a problem involving a lawn mower enchanted with magical powers. As thanks, Ludo got Arthur prime tickets to the 1994 Quidditch World Cup... \\ \hline
\cellcolor[gray]{0.8} \textbf{Fandom Description}\\ \hline
Otto Bagman was the brother of Ludovic Bagman. \textcolor{mygreen}{He once had a problem with a magical lawn mower, a Muggle artifact. Arthur Weasley helped him out with the problem, and was rewarded by Ludo with tickets to the 1994 Quidditch World Cup final.}\\ \hline
\cellcolor[gray]{0.8} \textbf{Human-authored Description}\\ \hline
Otto Bagman is the brother of Ludovic Bagman. \textcolor{mygreen}{He had a problem involving a lawn mower enchanted with magical powers. He was helped by Arthur and gave Arthur prime tickets to the 1994 Quidditch World Cup.}\\ \hline
\cellcolor[gray]{0.8} \textbf{BART}\\ \hline
Otto Bagman was a fictional character in the \textcolor{red}{1994} film Harry Potter.\\\hline
\cellcolor[gray]{0.8} \textbf{T5-base}\\ \hline
Otto Bagman \textcolor{red}{was an English footballer who played for the Wimbourne Wasps and the English National Quidditch team. He also played dirty when gambling and betting as he tried to find loopholes or even pay in fake money.} \\\hline
\cellcolor[gray]{0.8} \textbf{T5-large}\\ \hline
Otto Bagman was a brother of Ludovic Bagman. \\ \hline
\end{tabular}
\caption{Examples of entity descriptions generated by our model. Red text indicates incorrect information in predictions while green text indicates information in the Wikipedia and human-authored descriptions that was not covered by any of the model predictions.}  \label{tbl:example}
\end{table*}

\end{document}